\documentclass[10pt, conference, compsocconf]{IEEEtran}

\ifCLASSINFOpdf
\else
\fi
\usepackage{algorithmic}
\usepackage{array}
\usepackage{graphicx}
\usepackage{diagbox}
\usepackage{multirow}
\usepackage{subfigure}
\usepackage{amsmath}
\usepackage{amssymb}
\usepackage[numbers,sort&compress]{natbib}

\nocite{*}
\newcommand{\topcaption}{%
\setlength{\abovecaptionskip}{0pt}%
\setlength{\belowcaptionskip}{10pt}%
\caption}

\begin{document}

%
\title{PGN: A perturbation generation network against deep reinforcement learning}

\author{  
    \IEEEauthorblockN{Xiangjuan Li\IEEEauthorrefmark{1}, Feifan Li\IEEEauthorrefmark{1}\IEEEauthorrefmark{2}, Yang Li\IEEEauthorrefmark{1} and Quan pan\IEEEauthorrefmark{1}}
    \IEEEauthorblockA{  
        \IEEEauthorrefmark{1}Northwestern Polytechnical University, Xi'an, China\\  
        lixiangjuan@mail.nwpu.edu.cn, \{liff, liyangnpu, quanpan\}@nwpu.edu.cn
}}



%


\maketitle

\begin{abstract}
Deep reinforcement learning has advanced greatly and applied in many areas. In this paper, we explore the vulnerability of deep reinforcement learning by proposing a novel generative model for creating effective adversarial examples to attack the agent. Our proposed model can achieve both targeted attacks and untargeted attacks. Considering the specificity of deep reinforcement learning, we propose the action consistency ratio as a measure of stealthiness, and a new measurement index of effectiveness and stealthiness. Experiment results show that our method can ensure the effectiveness and stealthiness of attack compared with other algorithms. Moreover, our methods are considerably faster and thus can achieve rapid and efficient verification of the vulnerability of deep reinforcement learning.

\end{abstract}


%
\IEEEpeerreviewmaketitle

\section{Introduction}
In recent years, Deep Reinforcement Learning (DRL) has achieved excellent results in many application areas such as games\cite{lample2017playing}, robot control\cite{tai2017virtual}, and other tasks\cite{miao2022research,peng2021phonetic,DBLP:journals/chinaf/NingS00W0K21}, and even surpassed human levels. However, as a combination of deep learning and reinforcement learning, DRL inevitably has vulnerabilities, especially in deep learning, where images face a variety of complex attack forms, such as the Fast Gradient Sign Method (FGSM)\cite{goodfellow2014explaining}, Projected Gradient Descent (PGD)\cite{madry2017towards}, Carlini \& Wagner (CW)\cite{carlini2017towards} and Few Pixel Attack\cite{li2023few}. These cause the application of DRL is vulnerable to adversarial attacks, which can render the system ineffective or even catastrophic in critical scenarios such as autonomous driving. Thus, it's necessary to verify the vulnerability of deep reinforcement learning in the form of an adversarial attack in advance to prevent it from causing irreparable losses in the application.

In DRL, the agent's observations are generally presented in the form of images, so image attack methods can be used to tamper with the agent's observations, resulting in incorrect decision-making. There are many works about how to verify the vulnerability of the attack. Adversarial attacks against DRL were first introduced in 2015 by Huang et al.\cite{huang2017adversarial} and Kos et al.\cite{kos2017delving}, who adapted FGSM from image classification to the DRL. Then Lin et al.\cite{lin2017tactics} and Sun et al.\cite{sun2020stealthy} focused on how to reduce the attack frequency and attack only at critical moments. Li et al.\cite{li2022deep} investigated how to use an adversarial strategy network to adaptively select the moment of attack and generate an adversarial example for the attack, considering both the long-term reward and short-term reward. Although the aforementioned algorithms reduced the number of attacks, they still used the traditional method to produce the adversarial examples, which has the problems of high time complexity and low quality of the generated example images to verify the vulnerability quickly and efficiently. In addition, \cite{DBLP:journals/corr/abs-2301-07487} pointed out that the effect of attacking against the observation of an agent using traditional image transformation operations is much greater than that of traditional attack methods such as CW. Therefore, it is necessary to research how to produce adversarial examples against DRL to verify the vulnerability as fast as possible and insidiously. In the meantime, there is still a lack of an efficient evaluation metric for attacking stealthiness. 

Based on the above research, in this paper, we design a perturbation generation network (PGN) for observation-based attacks against DRL. The network aims to generate perturbation that makes the agent choose the wrong action and eventually leads to the failure of the task. The trained network can generate perturbations adaptively for the observations of the agent at different moments. Compared with traditional adversarial example generation algorithms, the adversarial examples generated by PGN make the agent less detectable, while ensuring the effectiveness of the attack and the quality of the adversarial examples, balancing the effectiveness and stealthiness of the attack. The key contributions of this paper are as follows:
\begin{itemize}
    \item To address the shortcomings of traditional attacks with high time complexity, a perturbation generation network is designed to generate perturbations quickly and efficiently to perturb the observation space of an agent in real-time.
    \item Combined with the characteristics of the attack against DRL, the perturbation network is constructed by introducing \textit{Q} value and can achieve targeted and untargeted attacks.
    \item The ratio of consistent actions without and with an attack is introduced to measure the stealthiness of the attack.
    \item A new attack effect measurement index is proposed by comprehensively consider the effectiveness and stealthiness of attack.
\end{itemize}

The rest of the paper is organized as follows. Section II introduces related works. Section III describes the proposed method. Section IV presents the details of the experiment and analyse the results. Finally, Section V concludes the paper and draws the future work.

\section{Related Works}
\subsection{Deep reinforcement learning}
Deep reinforcement learning enables agents to learn by interacting with the environment to achieve long-term rewards. The agent's interaction with the environment can be represented as Markov Decision Process(MDP). In general, it can be interpreted by a tuple $<S,A,P,r,\gamma>$, where $S$ is the state space, $A$ is the action space, $P$:$S\times A\times S \rightarrow [0$:$1]$ is the transition probability, $r$ denotes the reward, and $\gamma\in[0,1) $ is the discount factor. At each time step, the current state of the agent can be represented as $s_t$, and then it will select action $a_t$ to enter the next state $s_{t+1}$ according to the policy network $\pi$, where the policy $\pi(s_t,a_t)\in[0,1]$ represents the possibility of choosing an action and can also directly output the optimal action in the state. $r(s_t,a_t)$ represents the immediate reward after executing the action $a_t$. The goal of the agent is to learn an optimal strategy to maximize the final reward $R$, which can be interpreted as:
\begin{equation}
R=\sum_t^{T-1}E_{\pi(s_t,a_t)}[\gamma^tr(s_t,a_t)] 
\end{equation}

Generally, DRL can be categorized into value-based and policy gradient-based. The former primarily employs deep neural networks to approximate the target action-value function, which represents the cumulative reward obtained by reaching a certain state or performing a certain action. The representative methods such as Deep Q-Network (DQN)\cite{mnih2013playing}. 
 The latter\cite{schulman2015trust,lillicrap2015continuous}, on the other hand, parameterizes the policy and employs deep neural networks to approximate it while seeking the optimal policy along the direction of policy gradients.

\subsection{Adversarial attacks against machine learning}
Adversarial attacks in the context of deep reinforcement learning draw inspiration from the concept of adversarial attacks against machine learning models and can be considered a subcategory of such attacks.
Adversarial attacks in machine learning refer to the technique of deceiving machine learning systems into producing incorrect outputs or decisions when exposed to adversarial examples. Adversarial examples are modified versions of the original examples that are artificially created by adding, deleting, or changing features or noises. These examples can be generated through methods like  gradient-based methods\cite{goodfellow2014explaining,madry2017towards}, optimization methods\cite{carlini2017towards} and generative adversarial methods \cite{goodfellow2020generative,GargTV2020,jiawei2023conditional}.

According to the target, adversarial attacks can be classified into two types: targeted and untargeted. The former aims to classify an image into a designated category. The latter aims to make the model produce incorrect decisions without any specific goal.
Adversarial attacks can also be classified into two types based on the level of information available to the attacker: clear box and opaque box. In a clear box attack, the attacker has complete access to the model information, including its structure and parameters. In contrast, an opaque box attack limits the attacker's access to only the model’s input and output.

Common adversarial attack methods pose a significant challenge to the robustness and reliability of machine learning by carefully crafting the small perturbations to mislead the model.
However, adversarial attacks toward deep reinforcement learning follow a similar principle but are tailored to the specific setting of deep reinforcement learning agents. 
In summary, adversarial attacks in the domain of deep reinforcement learning are rooted in the broader concept of adversarial attacks against machine learning models. However, they address the unique challenges posed by reinforcement learning settings and pose a crucial area of research for improving the robustness and security of deep reinforcement learning agents.

\subsection{Adversarial Attacks against DRL}
As DRL is increasingly applied to decision-making tasks requiring high security, researchers are conducting more investigations into its vulnerability using adversarial attacks. Adversarial attacks in DRL can be categorized into attack based reward\cite{zhang2020adaptive}, attack based policies\cite{behzadan2019adversarial}, attack based observations\cite{lin2017tactics,li2022deep,DBLP:journals/compsec/LiLFWP23}, attack based environment\cite{bai2018adversarial}, and attack based actions\cite{lee2020spatiotemporally}. The objectives of the above attacks are all aimed at causing the agent to make incorrect decisions and resulting in the minimization of cumulative reward.

Currently, adversarial attacks based on observations in DRL mainly rely on carefully selecting the attack moment and generating adversarial examples. Huang et al.\cite{huang2017adversarial} injected attacks at each time step, which reduced the performance of the agent, but made it easy for the attacker to observe the agent's actions going beyond acceptable limits and detect the attack. Later attacks focused on fewer specific time steps to ensure stealthiness. In the time strategy attack proposed by Lin et al.\cite{lin2017tactics}, they calculated the difference between the action probabilities for each step and launched attacks when the difference exceeded a given threshold, which reduced the number of attacks to some extent. Li et al.\cite{li2022deep} studied how to use an adaptive adversarial policy network to select attack moments and generate adversarial examples for the attack while considering both the long-term reward and short-term reward of the agent. However, the above methods require either calculation of attack moment or a large amount of pre-training and still use traditional methods such as FGSM, CW, and PGD to generate adversarial examples, neglecting the time complexity required for example generation. When verifying the vulnerability of DRL algorithms, it is essential to reduce the time and computational costs as much as possible.

\section{Methodology}
In this paper, we mainly research the clear box attack against well-trained DQN agents to verify the vulnerability of DRL. We design a generative model PGN that can add appropriate perturbations to the agent’s observations, resulting in the final reward being reduced, while also ensuring the effectiveness and stealthiness of the attack. The structure of AutoEncoder-based PGN is in Figure 1. Given a noise $z$, PGN generates a $\delta$ and adds it to the original observation $x$, and then will be clipped to get $h(x)$, which will be passed to the agent and confuses it.
\begin{figure*}
    \centering
    \includegraphics[width=4.82in]{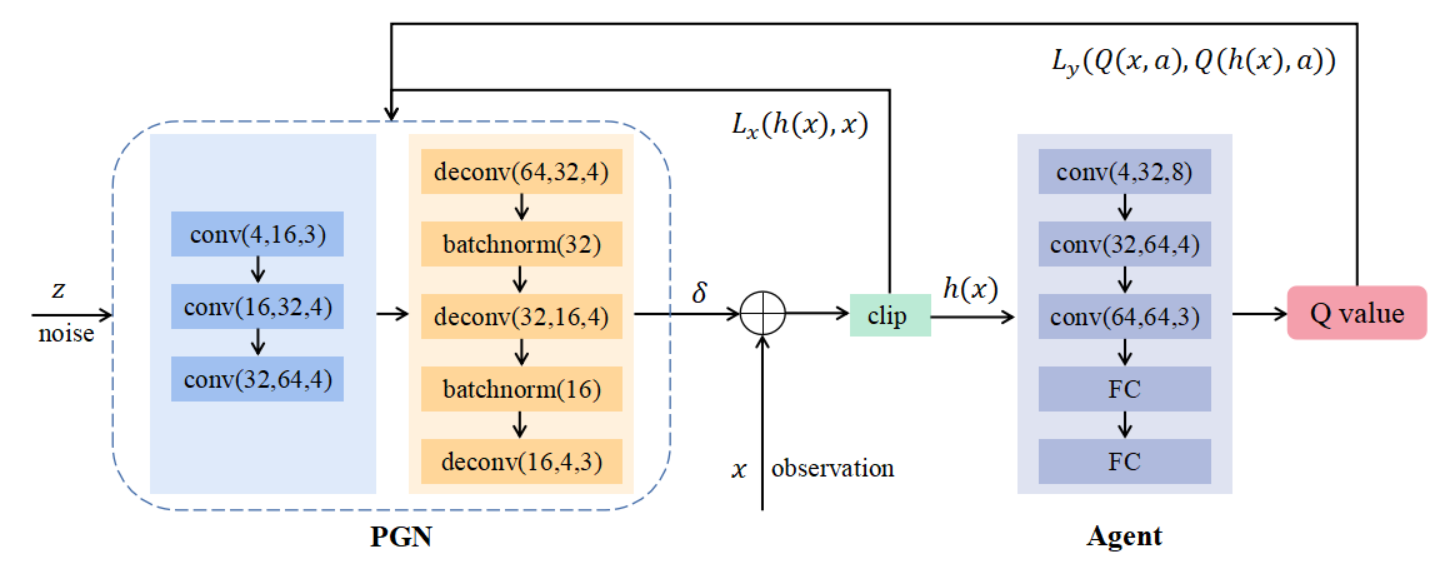}
    \caption{The structure of AutoEncoder-based PGN and training process}
    \label{fig2}
\end{figure*}

\subsection{Adversarial Network Model}
Given an agent model $f:X \rightarrow Y$, the observation of the agent can be represented as $x$, and the goal of the attacker is to find the perturbation $\delta$ and get the adversarial example $x+\delta$, which satisfies $f(x)\neq f(x+\delta)$. The PGN can be described as: $h(x)=x+\delta$, and that satisfies constraint in DQN:
\begin{equation}
\arg\max_{a}Q(x,a)\neq \arg\max_{a}Q(h(x),a)
\end{equation}
To ensure effectiveness, the ideal goal of PGN is to find a $\delta$ that also satisfies:
\begin{equation}
h(x)=\arg\max_{x+\delta}Q(x+\delta,\arg\min_{a}Q(x,a))
\end{equation}
and $\|\delta\|_2<\epsilon$ for $\epsilon>0$.
To achieve the above objective, the loss function of PGN can be designed as:
\begin{equation}
L(\theta)=\alpha L_x(x,h(x))+L_y(y,y')+\beta L_c
\end{equation}
The former represents the Euclidean distance between the original observation and adversarial observation:
\begin{equation}
L_x(\mathbf{x}, \mathbf{h(x)}) = \sqrt{\sum_{i=1}^n (x_i - h(x)_i)^2}
\end{equation}
where $n$ represents the dimension, $x_i$ and $h(x)_i$ represent the component in dimension $i$ of original observation and adversarial example, by minimizing it can ensure the maximum similarity of original observation and adversarial observation. The second is the loss of effectiveness. Inspired by Generative Adversarial Networks (GANs), the effectiveness loss can be designed by the following two schemes:

Untargeted attack (U-PGN): We introduce $L_{y1}$ that maximizes the difference between the maximum \textit{Q} value $Q(x,a_t)$ under normal observations and the \textit{Q} value under adversarial observation and related action $a_t$. This encourages the agent to select actions other than the optimal one, similar to non-targeted attacks, in order to disrupt the agent's decision-making:
\begin{equation}
L_{y1}=10-\log L(Q_{\max}(x,a_t),Q(h(x),{\arg\max}_a Q(x,a)))
\end{equation}

Targeted attack (T-PGN): to ensure the effectiveness of the attack, the target action is defined as the action corresponding to the minimum \textit{Q} value under the original observation. As a result, the difference between the \textit{Q} values of the target action under original and adversarial observations should be maximized, thereby ensuring that the \textit{Q} value of the target action under adversarial observation is maximized as much as possible, and the probability of it being selected is also maximized:
\begin{equation}
L_{y2}=L(Q(h(x),a_{targeted}),\mathbf{R}(Q(x),a_{targeted}))
\end{equation}
\begin{equation}
a_{targeted}={\arg\min}_a Q(x,a)
\end{equation}
in which $\mathbf{R(\cdot)}$ is the re-ranking strategy.

All of the two schemes are calculated by Mean Squared Error (MSE) as follows:
\begin{equation}
 L_y = \frac{1}{N}\sum_{i=1}^{N}(y_i - \hat{y}_i)^2   
\end{equation}
where $n$ represents the dimension, $y_i$ and $\hat{y}_i$ represents the component in dimension $i$.

At last, we introduced the hinge loss term by comparing the $L_2$ norm to constrain the perturbation within a certain range, in which the $C$ means the threshold of perturbation:
\begin{equation}
L_c=\max(\|\delta\|_2 - C,0)
\end{equation}
we will penalize the PGN if the $L_2$ norm of perturbation is greater than $C$.

\subsection{Re-ranking Strategy}
In order to achieve a strong-targeted attack, in this work we introduced the re-ranking strategy inspired by ATN\cite{baluja2018learning}. Especially, the re-ranking strategy $\mathbf{R(\cdot)}$ can magnify the \textit{Q} value $Q(s_t, a_{targeted})$ of the target action $a_{targeted}$ and maintain the order of other actions. In this case, the perturbation can be ensured small enough and relevant only to the target action. The re-ranking strategy can be described as:
\begin{equation}
\mathbf{R}(Q,a_{targeted}) = norm\left(\left\{ \begin{array}{ll}
\kappa \cdot \max Q & \text{if }a_t=a_{targeted}\\
Q(s_t,a_t) & \text{otherwise}
\end{array} \right\}\right)
\end{equation}

Given the \textit{Q} value of the original observation, the action corresponding to the minimum \textit{Q} value is taken as the target action. Then the $Q(s_t,a_{targeted})$ will be changed to $\kappa$ times the maximum \textit{Q} value, and $\kappa > 1$ is an additional parameter specifying how much larger the targeted \textit{Q} value should be than the original max \textit{Q} value. $norm(\cdot)$ is a normalization function that rescales its input to be a valid probability distribution.

\subsection{Action Consistency Ratio}
In adversarial attacks against DRL, attackers aim to manipulate the agent's observations to make it choose the wrong actions, leading to the failure of the task. To ensure the stealthiness of the attack, attackers generally choose to inject attacks in as few time steps as possible to avoid being discovered. So it is essential to maintain a high level of consistency in the agent's action selection, which is known as the action consistency ratio. The goal is to make the attack virtually undetectable to bystanders while reducing the reward obtained by the agent. From a bystander's perspective, it is often difficult to determine whether an agent's observations have been tampered with, but can only judge whether the agent is attacked according to its actions. 

For instance, in the game of Pong, an agent typically moves toward the ball. If an agent's actions are always inconsistent with human expectations, it is highly likely that the agent has been attacked. When the ball is downward, the agent should also downward, but after the attack the agent chooses upward and away from the ball. If this happens frequently, then we consider the agent to be under attack. Therefore, we introduce the action consistency ratio $ACR$ to represent the stealthiness of the attack. If the agent's actions remain consistent with what is expected in most cases but the reward is ultimately reduced, then the stealthiness of the attack is guaranteed. The formula is expressed as:
\begin{equation}
ACR=N_{same}/N_{total}
\end{equation}
The $N_{same}$ means the number of actions that are consistent with both normal and attacked observations, and the $N_{total}$ means the total number of actions.

\section{Experiment}
\subsection{Experiment Settings}
The purpose of PGN is to effectively attack a normal agent in a real-time scenario. To achieve this, it is assumed that the attacker has access to the policy network of the agent. By predicting the action distribution of the agent before and after adding perturbations and using the difference in \textit{Q} value distribution as one of the loss terms, the model can be further adjusted.

In the experiment, four DQN agents corresponding to the Pong, MsPacman, SpaceInvaders, and Qbert games were pre-trained using PyTorch, Tianshou\footnote{Tianshou is a PyTorch-based reinforcement learning framework designed to provide efficient implementation and easy-to-use API.} and Gym\footnote{Gym is a Python open-source library designed for reinforcement learning which provides standardized interfaces and environments.} libraries. Then PGN utilizes the interaction of pre-trained agents with the environment to construct an offline dataset for training. The parameters of DQN and PGN are recorded in Table I and Table II respectively. Generally, the corresponding reward achieved by the four agents is shown in the ``Normal'' row of Table III.

We compared the PGN with CW, FGSM, and PGD under the same attack frequency which is 100\%. The $\epsilon$ in FGSM is 0.1, and the iterations of the CW and PGD are 50. PGN is designed with two different structures: AutoEncoder-based (as shown in Figure 1) and Generator-based (the structure of PGN is similar to the generator in GANs). Thus the targeted attack T-PGN can be represented as ``T-PGNA'' and ``T-PGNG'', the untargeted attack U-PGN can be represented as ``U-PGNA'' and ``U-PGNG''.
\begin{table}[htbp]
\topcaption{hyper-parameters in DQN}
\centering
\setlength{\tabcolsep}{2mm}{
    \begin{tabular}{c|c|c|c}
    \hline
         parameter &value &parameter &value \\ 
    \hline   
    learning rate & 1e-3 & batch size & 32\\
    \hline
    discount factor $\gamma$ & 0.99 & target network update frequency& 500\\
    \hline
    replay buffer & 1e5 & training episode & 16 \\
    \hline
    $\epsilon$-greedy start & 1 & $\epsilon$-greedy end & 0.05 \\
    \hline
\end{tabular}}
\label{tab1}
\end{table}

\begin{table}[htbp]
\topcaption{hyper-parameters in PGN}
\centering
\setlength{\tabcolsep}{4mm}{
    \begin{tabular}{c|c|c|c}
    \hline
         parameter &value &parameter &value \\ 
    \hline   
    learning rate & 1e-3 & batch size & 128\\
    \hline
    epoch & 40 & training episode & 20\\
    \hline
    $\alpha$ & 1e-2 & $\beta$ & 1 \\
    \hline
    $\kappa$ & 10 & $C$ & 0.1 \\
    \hline
\end{tabular}}
\label{tab2}
\end{table}

\subsection{Results}
The cumulative reward and $ACR$ of different attack methods in each game are shown in Table III. Each result is an average of over 10 episodes. In Table III, ``Normal'' represents the reward without attack, ``T-PGNA'' and ``T-PGNG'' represents the targeted attack of PGN, while ``U-PGNA'' and ``U-PGNG'' represents the targeted attack of PGN, and the bold data is the best result. 

\begin{table*}[htbp]
\topcaption{reward and $ACR$ of four attack methods}
\centering
\setlength{\tabcolsep}{4mm}{
    \begin{tabular}{c|cc|cc|cc|cc}
    \hline
     \multirow{2}{1cm}{Methods} & \multicolumn{2}{c|}{Pong}&\multicolumn{2}{c|}{MsPacman}&\multicolumn{2}{c|}{SpaceInvaders} &\multicolumn{2}{c}{Qbert}  \\
 \cline{2-9}
   & reward  & $ACR$ & reward & $ACR$ & reward & $ACR$ & reward & $ACR$ \\
    \hline   
    Normal & 21.0 &-- & 2155.0 &-- & 719.0  &-- & 5460.0 &--\\
    CW & -17.5 & 0 & 982.0 & 0 & 465.5 &10.95\% & 520.0 & 0.11\% \\
    FGSM &-21.0 &0.33\%  & 519.0 & 0.02\% & 258.5 &1.15\% & 225.0 & 2.10\%  \\
    PGD &-21.0 & 6.62\%  & 222.0 & 0 & \textbf{179.0} & 0  & \textbf{35.0} & 0 \\
    T-PGNA & \textbf{-21.0} & 8.77\% &\textbf{210.0} & 0 & 285.0 &\textbf{32.59\%} &125.0 & 11.92\% \\
    T-PGNG & -20.9 & \textbf{13.95\%} &315.0 &\textbf{3.13\%} &182.5 &30.56\% &70.0 & \textbf{19.80\%} \\
    U-PGNA & -21.0 & 3.88\% &2144.0 &98.57\% &499.5 & 96.70\% & 4390.0 &97.73\% \\
    U-PGNG &13.6 & 34.25\% &1600.0 & 96.48\% &547.0 & 63.48\% & 4465.0 &76.77\% \\
    \hline
\end{tabular}}
\label{tab3}
\end{table*}

\begin{table}[htbp]
\topcaption{PSNR value of four attack methods}
\centering
\setlength{\tabcolsep}{3.2mm}{
    \begin{tabular}{c|c|c|c|c}
    \hline
    Methods  & Pong & MsPacman & SpaceInvaders & Qbert \\
    \hline   
    CW & 10.65 &11.38 &\textbf{24.15} & \textbf{14.94} \\
    FGSM & 6.06 &10.94 &18.01 &13.23 \\
    PGD & 7.16 &8.05 & 9.59 &9.17 \\
    T-PGNA & \textbf{15.86} &\textbf{16.29} &21.91 &12.46 \\
    T-PGNG & 8.01 &8.97 & 13.11 &9.23 \\
    U-PGNG &9.16 &9.55 &11.13 &9.08 \\
    U-PGNA &22.40 &16.06 &24.15 &18.36 \\
    \hline
\end{tabular}}
\label{tab4}
\end{table}

\begin{figure}
    \centering
    \includegraphics[width=3.2in]{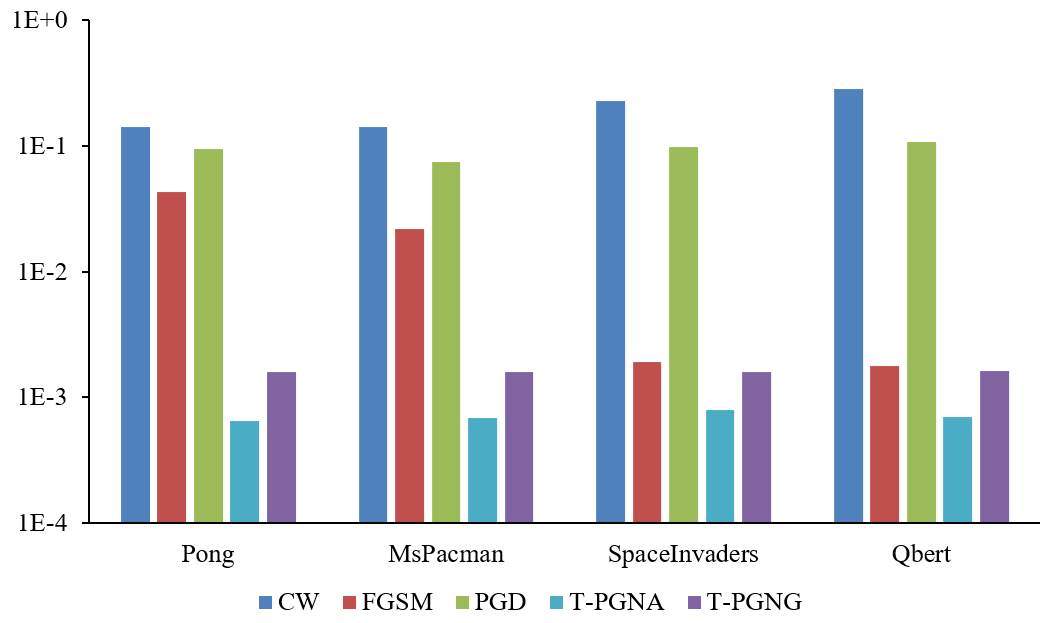}
    \caption{Time complexity of different attack methods}
    \label{fig3}
\end{figure} 

\begin{figure*}
    \centering
    \includegraphics[width=5.4in]{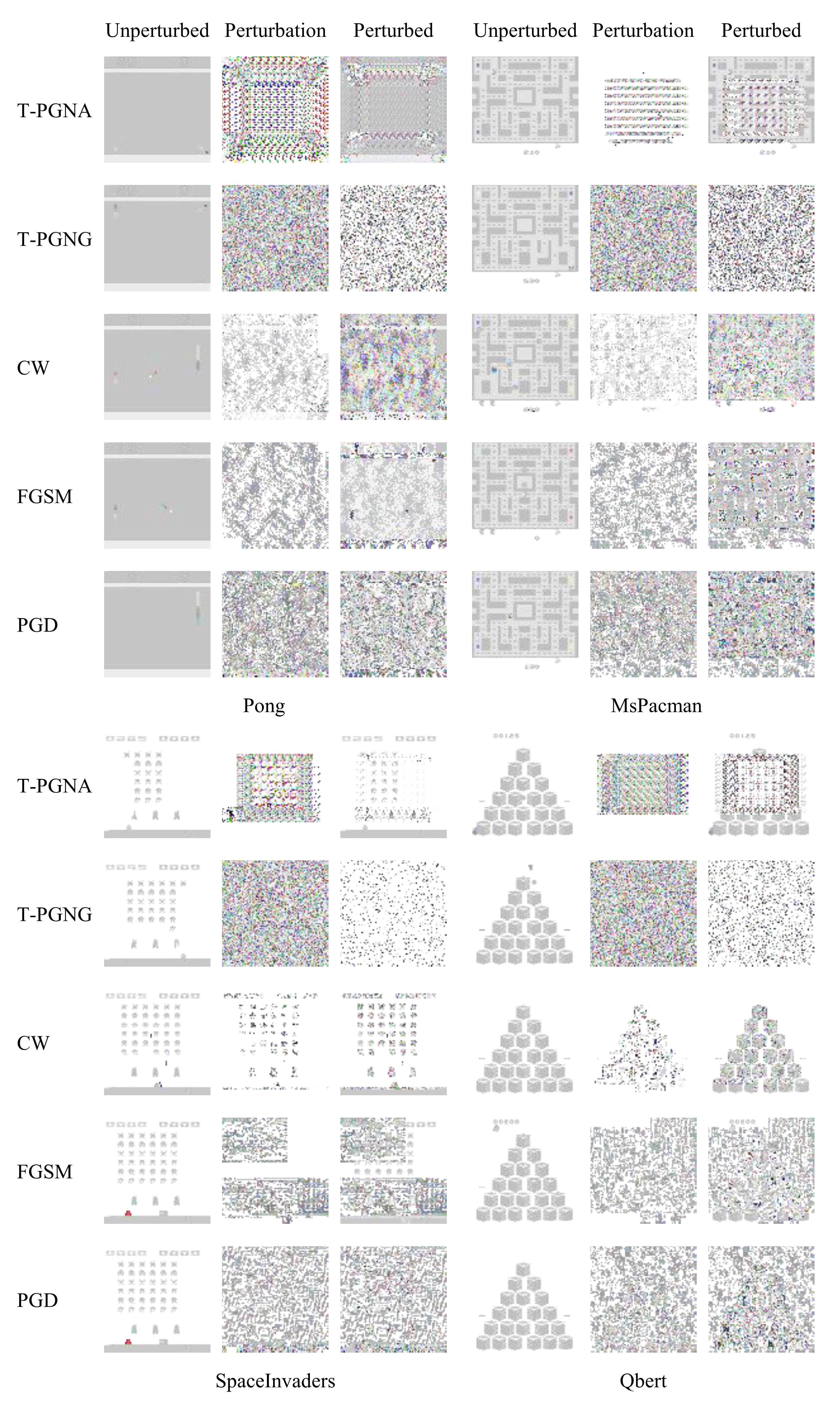}
    \caption{Unperturbed observations, perturbation, and perturbed observations of T-PGNA, T-PGNG, CW, FGSM, and PGD}
    \label{fig4}
\end{figure*}

And Table IV records the Peak Signal-to-Noise Ratio (PSNR) of different attack methods. The PSNR is used to measure the image quality of the adversarial example in an image attack. This index is based on the MSE of the image. The smaller the MSE, the larger PSNR, indicating better image quality. The PSNR value can be used to measure the image quality of the adversarial example. Specifically, the higher the PSNR value, the closer the adversarial example is to the original image; the lower the PSNR value, the worse the image quality.

If the $ACR>50\%$ and the percentage of reward reduced is less than 50\%, the attack is considered too weak. Therefore, according to the experimental results, U-PGNA and U-PGNG are weak attacks. We found that only U-PGNA performs the best in Pong, in the rest of the experiments both U-PGNG and U-PGNA have a low attack effect, but the $ACR$ is very high, which also shows that if the ratio of consistency of action with and without attack is very high it means the attack is almost ineffective. Thus we will not compare U-PGNA and U-PGNG in the following work.

From Table III we find that considering only the reward and PSNR, the lower the reward of the traditional attack method, the lower PSNR, i.e., the lower the image quality of the adversarial example. That is the attack effect at the expense of PSNR. In most cases, PGD has the lowest reward and the lowest PSNR, while CW has the highest reward and the highest PSNR in most cases. FGSM is in between. However, we should realize that different from traditional image classification attacks, attacks against DRL cannot be based solely on the quality of the adversarial examples as a measure of stealthiness. An attacker can easily determine whether an attack has suffered directly from changes in the agent's action. Therefore, the action consistency ratio $ACR$ is more important than PSNR in attack agents, and it is more meaningful to use it as a basis for judging stealthiness. 

Comparing the $ACR$, we find that T-PGNA and T-PGNG are always higher than traditional methods, even up to 32.59\% in SpaceInvaders, nearly 20\% higher than the other algorithms. However, the cumulative reward is slightly lower than PGD in several cases. It's an important question of how to balance the cumulative reward, $ACR$, and PSNR. Then we proposed a new index $AR$ for measuring the effectiveness and stealthiness based on the ratio of reward variation, $ACR$ and PSNR. The new index $AR$ can be expressed as:
\begin{equation}
AR=\alpha\times\Delta R + \beta\times ACR + \gamma\times \text{PSNR}
\end{equation}
\begin{equation}
\alpha + \beta + \gamma = 1
\end{equation}
where $\alpha$, $\beta$, and $\gamma$ are weighting coefficients, $\Delta R$ means the ratio of reward variation, the numerator is the difference between the cumulative reward of the normal agent and the minimum cumulative reward of the victim, and then the denominator is the difference between the cumulative reward of the normal agent and the victim, namely:
\begin{equation}
\Delta R=\frac{R_{normal}-R_{attacked}}{R_{normal}-\min{(R_{attacked}})}
\end{equation}
in which $R_{normal}$ represents normal reward of agent and $R_{attacked}$ represents reward after being attacked. According to different rules of games, the $\min{(R_{attacked})}$ value of Pong is -21, and 0 in the remaining three games. The first item of $AR$ represents effectiveness, and the last two represent the stealthiness of the attack. In calculation, the $\Delta R$ and the $ACR$ are in $0\sim 1$, PSNR is in $0\sim 100$, thus the weighting coefficients $\alpha$, $\beta$ and $\gamma$ are 0.5, 0.49 and 0.01 respectively.

\begin{table}[htbp]
\topcaption{$AR$ value of four attack methods}
\centering
\setlength{\tabcolsep}{3.2mm}{
    \begin{tabular}{c|c|c|c|c}
    \hline
    Methods  & Pong & MsPacman & SpaceInvaders & Qbert \\
    \hline   
    CW & 0.56 &0.39 &0.47 &0.60 \\
    FGSM &0.56 &0.49 &0.51 &0.62 \\
    PGD &0.60 &0.53 &0.47 &0.59 \\
    T-PGNA & \textbf{0.70} &\textbf{0.61} &\textbf{0.69} &0.67 \\
    T-PGNG &0.65 &0.53 &0.65 &\textbf{0.68} \\
    \hline
\end{tabular}}
\label{tab5}
\end{table}

Then we compared $AR$ in Table V. Results show that the $AR$ of T-PGNA and T-PGNG are in $0.6\sim0.7$, while others are lower than 0.6 in most cases. The results demonstrated that T-PGNA/G strikes a good balance between effectiveness and stealthiness compared to traditional attack methods.

Then, Figure 2 shows the time complexity of CW, FGSM, PGD, T-PGNA, and T-PGNG. We find that T-PGN is the fastest, with an order of magnitude around 1e-4 (T-PGNA) and 1e-3 (T-PGNG), FGSM is the next, and PGD is inferior to it, while CW is the slowest with an order of magnitude around 1e-1. The high time complexity is due to the fact that both PGD and CW are based on iterative optimization algorithms. This means that PGN can achieve fast and efficient verification of vulnerability in DRL, greatly reducing time costs. 

Considering the effectiveness of the attack, Figure 3 records the unperturbed observations, perturbation, and perturbed observations of T-PGNA, T-PGNG, CW, FGSM, and PGD in four games. The $L_2$ norm of each perturbation is limited to 0.1. As depicted in Figure 3, it is evident that the perturbations generated by T-PGNA and CW tend to be localized, particularly noticeable in SpaceInvaders and Qbert. Conversely, the perturbations produced by FGSM and PGD exhibit a more global influence. Simultaneously, the PSNR values for T-PGNA and CW significantly surpass those of FGSM and PGD. This disparity arises because FGSM and PGD employ gradient-based attack methods, optimizing perturbations by maximizing the loss function gradient. Consequently, these approaches affect the entire input image, resulting in global perturbations.

In contrast, CW follows a localized search strategy, striving to discover minimal perturbations within specific image regions to achieve adversarial objectives. It aims to identify regions with minimal perturbation that minimize the objective function. Furthermore, T-PGNA reconstructs input noise through encoding and subsequent decoding, ensuring maximum efficiency and minimal size.

\section{Conclusion}
In this paper, we explored how to verify the vulnerability of DRL quickly and efficiently by proposing the generative model PGN, which is based on GAN and AutoEncoder, ensuring the stealthiness and effectiveness of the attack. Meanwhile, considering the relationship between the attacker, the agent, and the environment, we proposed the action consistency ratio $ACR$ to measure the attack's stealthiness. Finally, we proposed an index $AR$ that can combine reward, $ACR$, and PSNR to measure the effectiveness and stealthiness of the attack. Experimental results show that compared to traditional attack methods FGSM, CW, and PGD, PGN can effectively reduce the agent's final reward while ensuring a higher $ACR$ and the time complexity is greatly reduced. The model in this paper can be used to verify the vulnerability of various DRL algorithms. However, in practical applications, algorithms with higher robustness tend to be directly selected instead of considering the vulnerability of the algorithm itself. Therefore, verifying the vulnerability of DRL is only the first step, and how to improve its robustness is the key. Therefore, in the following work, we will consider studying how to apply the detection method in \cite{dong2021cyber} on attack detection of DRL, and how to use regularization methods to further study the improvement of DRL robustness.

\section*{Acknowledgment}

This research is supported by the National Natural Science Foundation of China (No.62103330, 62203358, 62233014), the Fundamental Research Funds for the Central Universities (3102021ZDHQD09), and the Practice and Innovation Funds for Graduate Students of Northwestern Polytechnical University.



%

\bibliography{ref}

\end{document}